\documentclass[letterpaper,twocolumn,fleqn]{article} 

\usepackage{ist}
\usepackage{multirow}
\usepackage{adjustbox}
\usepackage{amssymb}
\usepackage{amsmath}
\usepackage{graphicx}
\usepackage{array}

\title{A Novel Dataset for Non-Destructive Inspection of Handwritten Documents}

\author{ Eleonora Breci, Luca Guarnera, Sebastiano Battiato; Department of Mathematics and Computer Science, University of Catania, Italy}

\date{} 

\begin{document} 

\maketitle 

\thispagestyle{empty} 


\begin{abstract}



Forensic handwriting examination is a branch of Forensic Science that aims to examine handwritten documents in order to properly define or hypothesize the manuscript's author. These analysis involves comparing two or more (digitized) documents through a comprehensive comparison of intrinsic local and global features. If a correlation exists and specific best practices are satisfied, then it will be possible to affirm that the documents under analysis were written by the same individual. 
The need to create sophisticated tools capable of extracting and comparing significant features has led to the development of cutting-edge software with almost entirely automated processes, improving the forensic examination of handwriting and achieving increasingly objective evaluations. This is made possible by algorithmic solutions based on purely mathematical concepts.
Machine Learning and Deep Learning models trained with specific datasets could turn out to be the key elements to best solve the task at hand. 
In this paper, we proposed a new and challenging dataset consisting of two subsets: the first consists of 21 documents written either by the classic ``pen and paper" approach (and later digitized) and directly acquired on common devices such as tablets; the second consists of 362 handwritten manuscripts by 124 different people, acquired following a specific pipeline. Our study pioneered a comparison between traditionally handwritten documents and those produced with digital tools (e.g., tablets).
Preliminary results on the proposed datasets show that 90\% classification accuracy can be achieved on the first subset (documents written on both paper and pen and later digitized and on tablets) and 96\% on the second portion of the data. The datasets are
available at https://iplab.dmi.unict.it/mfs/forensic-handwriting-analysis/novel-dataset-2023/.
\end{abstract}

\section{Introduction}
Handwritten document analysis is a specialized field within forensic science, dedicated to the meticulous examination of intrinsic features \cite{guarnera2017graphj,guarnera2018forensic,koppenhaver2007forensic,morris2020forensic} in order to ascertain the authorship of documents. Its primary objective is to unveil the unique traits and patterns within handwritten materials, enabling the identification and attribution of authorship through a comprehensive investigative process. There are two methods of manuscript analysis: \textit{destructive inspection} and \textit{non-destructive inspection}.
\textit{Destructive inspection }involves all techniques that require the removal of a part of the writing. This investigation heavily depends on the type of document preservation, as exposure to sources of heat or humidity; it can degrade the document much more rapidly than normal. \textit{Non-destructive inspection} is used when it is necessary to preserve the original manuscript and can be carried out through the use of imaging techniques. 

\begin{figure}[t!]
    \centering     \includegraphics[width=\linewidth]{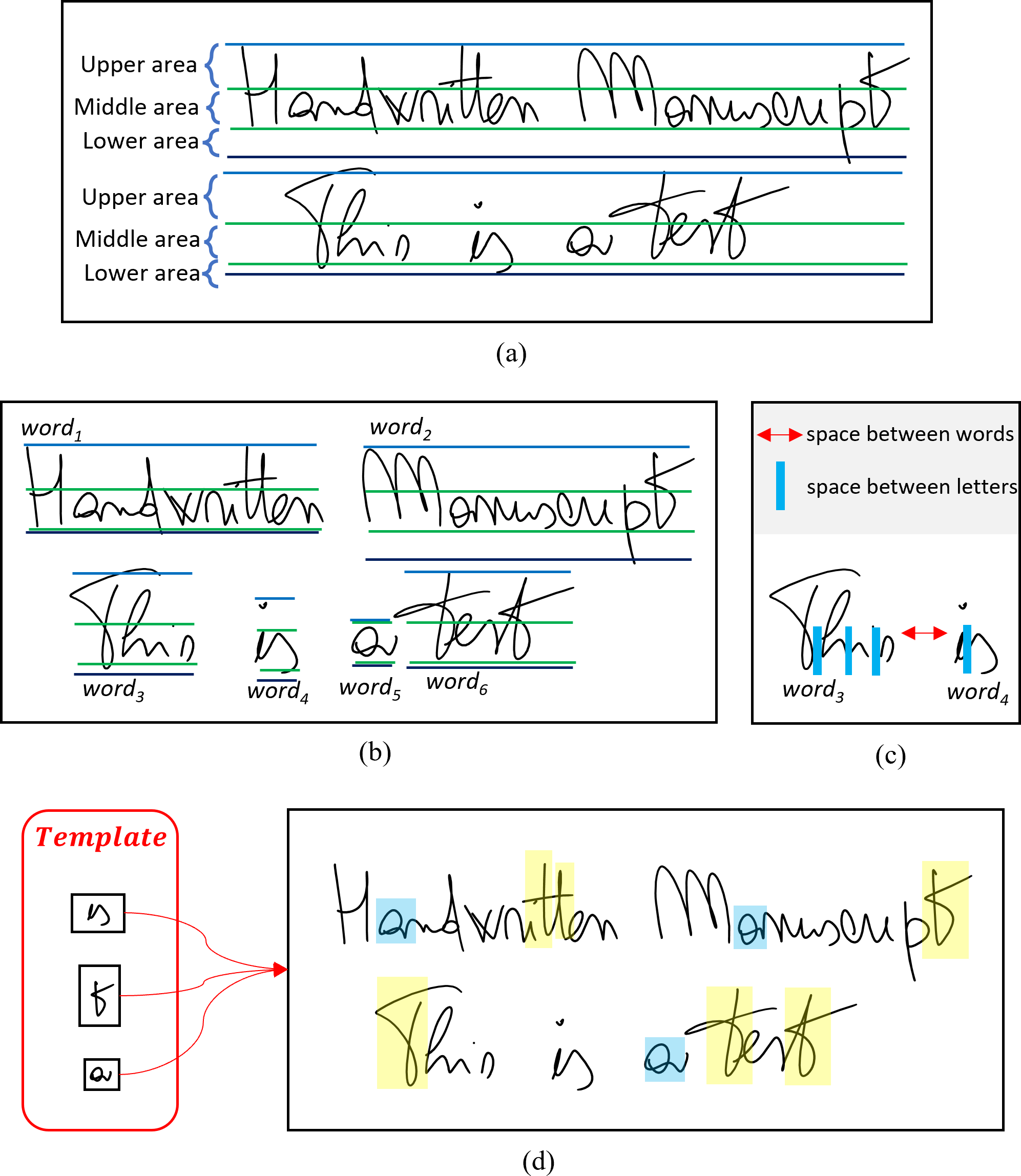}
    \caption{(a) Upper, middle and lower area detection; (b) Word detection; (c) Word and letter spacing; (d) Character recognition.}
\label{fig:textdescription}
\end{figure}

Law enforcement agencies use standard protocols based on manual processing of handwritten documents. This methodology is time-consuming, is often subjective in its evaluation, and is not fully replicable. To overcome these limitations, it is important to design new sophisticated software, based mainly on Deep Learning paradigms and techniques \cite{fiel2015writer,he2019deep,tang2016text}. The absence of specific and sophisticated datasets could severely affect the performance of algorithms based on artificial intelligence solutions. Therefore, in this paper, we present a new and challenging dataset consisting of 362 handwritten manuscripts by 124 different people, compiled following a specific scheme and rules. 
Each involved person wrote 3 documents: 
the first is a copy of a text projected on the screen; the second is a copy of a document written by an anonymous person, with the goal of ``copying/cloning" as much as possible calligraphy; and the last is a document containing a few sentences of a piece of music chosen at will. This approach to dataset creation is different from all other existing ones (such as \cite{crawford2020database,kleber2013cvl}), which instead include multiple texts written by the same author but all differing from each other.
In addition, 21 more images written by different people were included, using not only the classic ``pen and paper" approach and later digitized but also written directly on digital devices such as tablets.


A preliminary analysis of the proposed dataset was carried out by using the method \cite{breci2023innovative}. In detail, for each involved document, measures related to text line heights, space between words, and character sizes using image processing and deep learning techniques were extracted and processed. 
Each algorithm deals with a different aspect of text handling and retrieval, collectively contributing to a versatile set of tools for text analysis. The text line detection algorithm is designed to efficiently identify entire sentences from a given text document. In this phase
all the discriminative measurements are extracted: upper, middle and lower area (Figure \ref{fig:textdescription} (a)). 
Subsequently, the word detection algorithm was used to identify all the words (Figure \ref{fig:textdescription} (b)). 
We obtain also other discriminative measurements, such as: space between two words and the space between letters (Figure \ref{fig:textdescription} (c)). 
The third method, based on deep learning, requires the active involvement of a forensic expert who chooses a specific character, called a ``template", that will be searched in the entire document (Figure \ref{fig:textdescription} (d)).
Measurements such as its height and width are extracted as discriminative data points. 

For each involved document, the final feature vector consists of the mean ($\eta$) and standard deviation ($\sigma$) for every type of measure collected. In case the Euclidean distance between these features exceeds a predefined threshold, it is possible to affirm that the documents are written by the same person, otherwise by different authors. 
Experimental results on the proposed dataset show that with a classification accuracy of about 90\% on the first subset (documents written both on pen and paper and later digitized and on tablets) and 96\% on the second portion of the data, we can well define the authorship of a manuscript.
The obtained results could be improved by creating a more explainable and robust approach, considering the features computed through the Discrete Cosine Transform (DCT), which has been shown to achieve high results in resolution and explainability in several domains of image forensics, such as Deepfake detection.~\cite{giudice2021fighting,guarnera2023improving}. 


The main contributions of this paper are summarized below:
\begin{itemize}
    \item A new and challenging dataset consisting of 362 manuscripts written by 124 different people and another 21 manuscripts also written with digital devices was proposed. 
    \item 
    For the first time, forensic analysis of a manuscript incorporates digital documents, including those written on tablets.
    \item 
    Our dataset includes not only the original calligraphy of the authors but also how they attempt to mimic another calligraphy.

\end{itemize}

\section{Related Works}
\label{sec:sota}

Several studies~\cite{deviterne2020interpol,koppenhaver2007forensic,morris2020forensic} have analyzed the problem of writing identification with the aim of defining the most discriminating features to assess the identity of the writer. Morris~\cite{morris2020forensic} emphasized the importance of evaluating dimensional parameters by comparing absolute and relative quantities in order to analyze speed, slant, and style and detect possible attempts at forgery. According to Koppenhaver~\cite{koppenhaver2007forensic}, the analysis of character heights can provide discriminative information about an individual's handwriting variability. The importance of absolute dimensions in graphology has been studied by Hayes~\cite{hayes2006forensic}. These dimensions reflect hand and finger movements, which are influenced by individual characteristics. For example, some people tend to produce small handwriting, while others have larger handwriting.
Kelly and Lindblom~\cite{bisesi2006scientific} have shown that the relationship between the heights of upper and lower case letters can be used to identify the person who wrote a text.  
Guarnera et al.~\cite{guarnera2017graphj,guarnera2018forensic} showed that the relationship between certain intrinsic measures (e.g., distance between words and between letters, height and width of characters, etc.) can be used to identify the person who wrote a text.

Some groups of researchers have suggested the use of computational methods to perform text author recognition and verification.
Instead, Saad~\cite{saad2011application} designed a system using fuzzy logic and genetic algorithms to identify authors of Arabic handwritten texts offline by managing the ambiguity of handwriting similarities. 
Chahi et al.~\cite{chahi2018block} proposed the ``blockwise local binary count" operator to characterize each author's writing style by computing a set of histograms from the connected components detected in handwriting. This method allows authors to be identified in handwritten documents. 
Shivram et al.~\cite{shivram2013hierarchical} introduced a theoretical 
framework based on latent Dirichlet allocation to model writing styles as a shared component of an individual's handwriting. 
To identify the writer, Hannad et al.~\cite{hannad2016writer} proposed a method based on splitting the writing into small fragments. Each fragment is represented using texture-based descriptors such as histograms of local binary patterns, local ternary patterns and local phase quantization. The extracted descriptors are used to compare two documents by calculating the distance between them.
Several handwriting analysis tasks beyond author identification have also been explored. Bhardwaj et al.~\cite{bhardwaj2010retrieving} defined a method that can estimate similar handwriting styles corresponding to a given query image. In contrast, Ramaiah et al.~\cite{ramaiah2012handwritten} estimated the approximate age of historical handwritten documents by learning a distribution of different styles across centuries. 


Recent state-of-the-art approaches have addressed the task of identifying authors using machine learning and deep learning techniques on different datasets. Crawford et al.~\cite{crawford2023statistical} presented a statistical approach to model various handwriting styles, aiming to calculate the likelihood of a questioned document's authorship from a known set of writers. 
Semma et al.~\cite{semma2021writer} extracted key points from handwriting and using Convolutional Neural Networks (CNNs) for classification. FAST key points and Harris corner detectors identify points of interest, and a CNN is trained on patches around these points. Bennour et al.~\cite{bennour2019handwriting} explored writer characterization in handwriting recognition using an implicit shape codebook technique. Key points in handwriting are identified and used to create a codebook, leading to promising results in various experimental scenarios. Kumar et al.~\cite{kumar2020segmentation} introduced a writer identification model (SEG-WI) using a CNN and weakly supervised region selection. It achieves segmentation-free writer identification on various datasets, outperforming state-of-the-art methods. Lai et al.~\cite{lai2020encoding} proposed novel techniques using pathlet and unidirectional SIFT features for fine-grained handwriting description. He et al.~\cite{he2020fragnet} presented FragNet, a deep neural network with two pathways to extract powerful features for writer identification from word and page images. Bahram~\cite{bahram2022texture} used co-occurrence features extracted from preprocessed regions of interest and contour texture.

\section{Proposed Dataset}
The proposed dataset comprises two distinct subsets. The initial subset comprises 21 documents created through the traditional ``pen and paper" method, subsequently digitized, or directly acquired using common devices like tablets. The second subset comprises 362 handwritten manuscripts contributed by 124 individuals, obtained through a defined pipeline. The subsequent sections provide additional information on the acquisition, organization, and compilation of the data.

\begin{figure}[t!]
    \centering     \includegraphics[width=\linewidth]{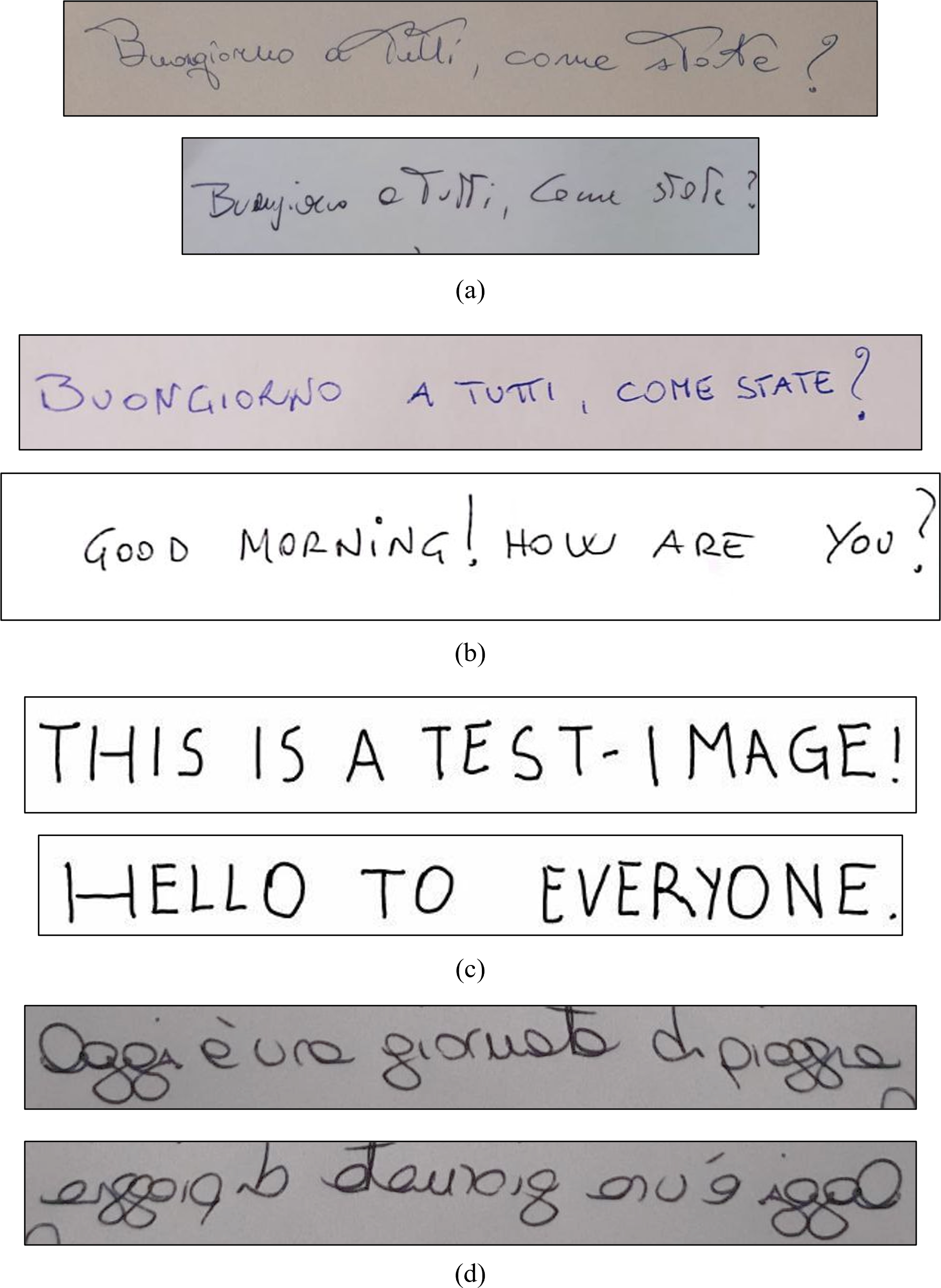}
    \caption{Examples of four different comparisons. (a) Two handwritten manuscripts written by different persons. (b) A handwritten manuscript is compared to a digitized text. (c) Two digitized texts written by the same person are compared. (d) Two identical manuscripts are compared, but one is mirrored.}
\label{fig:imagecompared}
\end{figure}

The first subset involved 21 manuscripts written by 13 different individuals. The peculiarity of this subset is its non-uniformity; each author has written their texts in their original handwriting style, language, using the medium they most frequently prefer, whether it be pen (without any distinction of colour) or digital devices (e.g. tablet). As a result, there are 21 distinct sentences with varying spontaneous styles (including bold or italics), all of which the proposed method can equally compare.
\begin{figure*}[t!]
    \centering     \includegraphics[width=\linewidth]{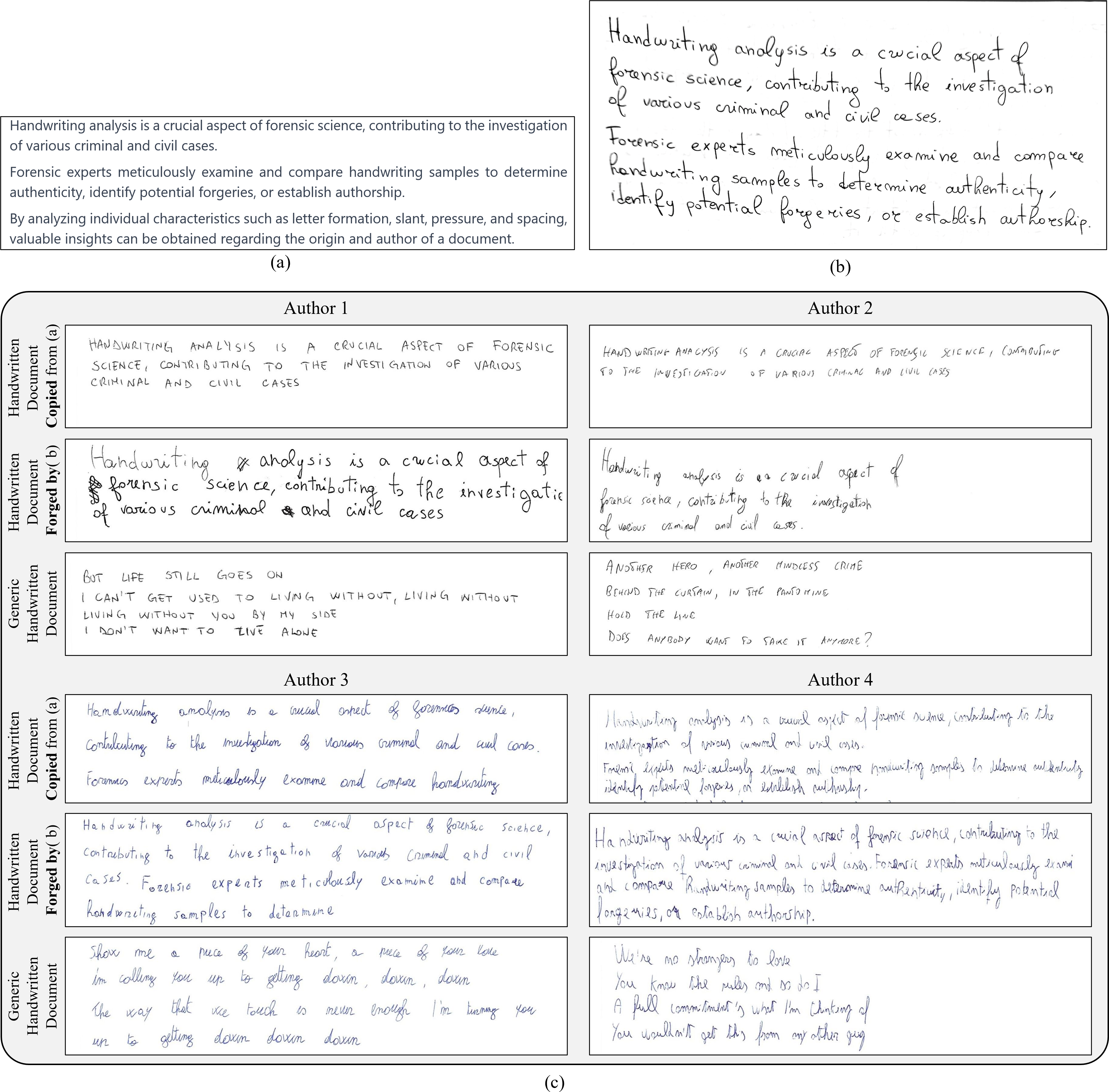}
    \caption{Some details of the proposed dataset. (a) Document projected and copied by the participants (with their original calligraphy); (b) The handwritten document to be forged (copied in the same calligraphy). (c) Some examples of documents written by 4 different people. In detail, for each person, the first row represents the handwritten document copied by (a); the second row represents the handwritten document forged by (b); finally, the third row represents a handwritten document with a randomly chosen text of a piece of music.}
\label{fig:dataset}
\end{figure*}
A preliminary analysis of this first portion of the proposed dataset was performed in order to define the authorship of the manuscripts, using the algorithms proposed in \cite{breci2023innovative}. In detail, all documents were compared with each other, resulting in a classification accuracy value (in the task of author identification by analyzing manuscript documents) of about 90\%. Table \ref{tab:comparison} shows some examples of the obtained results, by plotting 10 of the 169 examples of comparisons. 


\begin{table}[t!]
\centering
\begin{adjustbox}{max width=.35\textwidth}
\begin{tabular}{cccc}
\hline
\begin{tabular}[c]{@{}c@{}}\textbf{Experiment ID}\end{tabular} & \begin{tabular}[c]{@{}c@{}}\textbf{Euclidean} \\ \textbf{Distance}\end{tabular} & \begin{tabular}[c]{@{}c@{}}\textbf{Expected} \\ \textbf{Result}\end{tabular} & \begin{tabular}[c]{@{}c@{}}\textbf{Actual} \\ \textbf{Result}\end{tabular} \\ \hline
1                                                                        & 5,86                                                                   & D                                                                   & D                                                                 \\ \hline
2                                                                        & 9,72                                                                   & D                                                                   & D                                                                 \\ \hline
3                                                                        & 1,28                                                                   & S                                                                   & S                                                                 \\ \hline
4                                                                        & 11,88                                                                  & D                                                                   & D                                                                 \\ \hline
5                                                                        & 7,04                                                                   & D                                                                   & D                                                                 \\ \hline
6                                                                        & 5,44                                                                   & S                                                                   & D                                                                 \\ \hline
7                                                                        & 0,046                                                                  & S                                                                   & S                                                                 \\ \hline
8                                                                        & 7,05                                                                   & D                                                                   & D                                                                 \\ \hline
9                                                                        & 10,97                                                                  & S                                                                   & S                                                                 \\ \hline
10                                                                       & 1,42                                                                   & S                                                                   & S                                                                 \\ \hline
\end{tabular}
\end{adjustbox}
\caption{\textit{Table 1. }Examples of handwritten document comparisons. The first column describes the experiment ID number carried out, the second column shows the Euclidean distance resulting from the comparison of two feature vectors, the third column shows the result obtained (represented as ``D" for ``Different writer" and ``S" for ``Same writer"), and the fourth column shows the expected result.}
\label{tab:comparison}
\end{table}

The second portion of the proposed dataset includes manuscripts by various authors, organized and collected according to very specific rules and methods. The dataset is composed by 
362 images written by 124 students from the University of Catania, Italy. Each student wrote 3 manuscripts with the following rules: (i) to copy a computer-written projected text; (ii) to copy a text written in a sheet of paper with the main objective of reproducing as much as possible the same calligraphy; (iii) to write a few lines of a lyric of any song chosen by the user. 
Point (ii) turns out to be extremely important in the field of forensic investigations, mainly in cases where a malicious individual would attempt to simulate the calligraphy of any person for illicit purposes. This new dataset can then be extensively analyzed to create increasingly sophisticated algorithmic solutions capable of detecting forged documents of all kinds. 
In summary, the 3 documents written by each student consist of:
\begin{enumerate}
    \item The same text of the projected document with their original calligraphy.
    \item The same text of the projected document trying to copy the calligraphy of the writer.
    \item An arbitrary text with their original calligraphy. 
\end{enumerate}
Figure \ref{fig:dataset} shows the projected documents to be copied (for tasks 1 and 2) and some examples of the results obtained by 4 different students.
Each image was scanned and saved to test the \cite{breci2023innovative} method in defining authorship. The experimental results show a classification accuracy of $96\%$.



\section{Datasets in Comparison}

Several datasets used by researchers in this area are available in the literature such as:
\begin{itemize}
    \item \textbf{(i)} Computer Vision Lab (CVL) Database \cite{kleber2013cvl}, includes handwritten text from 311 unique writers, comprising both English and German samples. We selected a total of 1600 handwritten documents. 
    \item \textbf{(ii)} Center for Statistics and Applications in Forensic Evidence (CSAFE) Handwriting Database \cite{crawford2020database}, that consist of a total of 2430 handwriting sample images collected from surveys of 90 writers.
\end{itemize}

\begin{figure}[t!]
    \centering     \includegraphics[width=.8\linewidth]{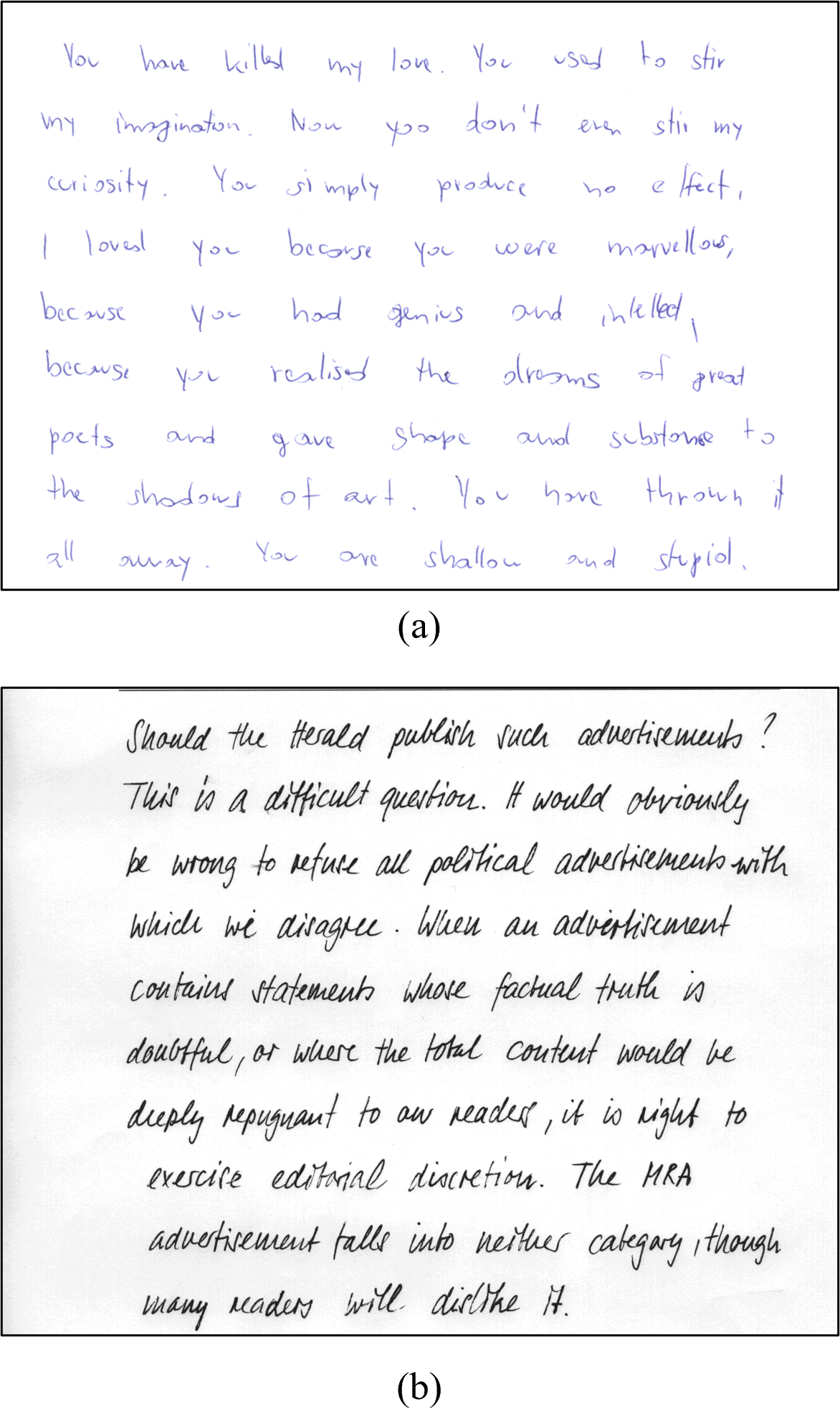}
    \caption{Examples of (a) CVL \cite{kleber2013cvl} and (b) CSAFE \cite{crawford2020database} digitized handwritten documents.}
\label{fig:cvl-iam-dataset}
\end{figure}

Figure \ref{fig:cvl-iam-dataset} shows some visual examples.

Our dataset sets itself apart from existing ones in the state of the art by including three documents from each author, capturing them in diverse scenarios. This unique aspect enables us, unlike other datasets, to compare each author's original handwriting on the same text. It also allows us to assess whether the proposed method can discern between authentic handwriting and attempts at imitation. Moreover, we can analyze distinct handwriting styles on various texts for each author. A significant distinction is that our dataset incorporates a blend of digitized documents
(crafted on tablets) and traditional pen-and-paper documents.

\section{Discussion, Conclusion and Future Works}

In conclusion, our study contributes significantly to the evolving landscape of Forensic Handwriting Examination by introducing a novel dataset that encompasses both traditional ``pen and paper" manuscripts and documents produced through contemporary digital devices like tablets. 

Our proposed dataset comprises two distinct subsets. The first subset involves 21 documents, initially written using the classic ``pen and paper" approach and later digitized, alongside documents directly created on common devices such as tablets. The second subset includes 362 handwritten manuscripts from 124 different individuals, acquired through a specific pipeline, allowing for a comprehensive exploration of both traditional and digital writing mediums.

Through a preliminary analysis on these datasets, we employed cutting-edge automatic algorithms for feature extraction, considering measures such as the heights of upper, middle, and lower areas of text lines and words, as well as the height and width of each character and the spaces between words, among others. To ascertain authorship, we calculated the Euclidean distance between feature vectors of different manuscripts. The classification accuracy achieved is noteworthy, standing at 90\% for the first subset (comprising documents written on pen and paper and later digitized, as well as on tablets) and an impressive 96\% for the second subset of the dataset.

This study not only showcases the effectiveness of the state of the art methodology but also highlights the importance of embracing both traditional and digital writing mediums in forensic handwriting examination. 

Looking ahead, our study opens avenues for future research and development in the field of Forensic Handwriting Examination. One key aspect that warrants attention is the expansion and enrichment of the proposed dataset, paving the way for more comprehensive and nuanced analyses. In general, the extension could include the following factors.

\begin{itemize}
    \item \textbf{Diverse Writing Styles and Contexts:} documents that reflect a broader spectrum of writing styles, contexts, and genres. This expansion would enhance the dataset's applicability to a wider range of real-world scenarios, providing a more realistic foundation for forensic examinations.
    \item \textbf{Temporal Evolution:} capture the temporal evolution of an individual's handwriting. Adding documents from various time periods in an individual's life would contribute to understanding how handwriting characteristics may change over time, influencing the accuracy and reliability of authorship determination.
    \item \textbf{Incorporation of Additional Digital Devices:} extend the dataset to include documents produced on a broader array of digital devices. The inclusion of handwriting samples from various tablets, stylus technologies, and other emerging digital platforms would further diversify the dataset and account for the evolving landscape of digital writing tools.
    \item \textbf{Increased Sample Size:} enlarge the dataset with a more extensive collection of handwritten manuscripts and documents. A larger sample size enhances statistical robustness and facilitates a more in-depth exploration of the relationships between extracted features, contributing to the overall efficacy of forensic analyses.
    \item \textbf{Annotation for Ground Truth:} annotate the dataset with additional ground truth information. Including annotations that identify specific characteristics or attributes within the documents can provide valuable insights for algorithmic training and evaluation.
    \item \textbf{Integration of Multimodal Data:} Explore the integration of multimodal data, such as incorporating handwriting samples alongside other biometric or contextual information. This holistic approach could lead to more comprehensive forensic examinations that consider a broader range of factors influencing authorship determination.
\end{itemize}

By addressing these considerations, future research can build upon the foundation laid by this study, advancing the capabilities of forensic handwriting analysis and ensuring that the proposed methodologies remain robust and applicable in evolving forensic contexts.

\section{Acknowledgments} 
This research is supported by Azione IV.4 - ``Dottorati e contratti di ricerca su  tematiche dell’innovazione" del nuovo Asse IV del PON Ricerca e Innovazione 2014-2020 “Istruzione e ricerca  per il recupero - REACT-EU”- CUP: E65F21002580005.

\small
\bibliographystyle{fullname}
\bibliography{ref}




\small

\begin{biography}

Eleonora Breci was born in Lentini on August 22, 2000. She completed her bachelor's degree in Computer Engineering at the University of Catania, discussing the thesis entitled ``GraphJ-3.0: Graphometric Measures for the Forensic Examination of Handwritten Documents" on October 7, 2022. Currrently, she is pursuing a Master's program in Cybersecurity at La Sapienza University, Rome. Her primary research areas focus on the forensic study of handwritten documents, with a particular attention on document authentication.

Luca Guarnera was born in Catania on October 26, 1992. Since January 1, 2022, he is a research fellow in Computer Science at the University of Catania. He graduated as Ph.D. in Computer Science (XXXIII cycle, PON number E37H18000330006) on October 14, 2021, discussing the thesis entitled “Discovering Fingerprints for Deepfake Detection and Multimedia-Enhanced Forensic Investigations” at the Department of Mathematics and Computer Science, University of Catania.  His main research interests are Computer Vision, Machine Learning, Multimedia Forensics and its related fields with a focus on the Deepfake phenomenon.

Sebastiano Battiato is a full professor of Computer Science at the University of Catania. He received his degree in Computer Science (summa cum laude) in 1995 from the University of Catania and his Ph.D. in Computer Science and Applied Mathematics from the University of Naples in 1999. He has been Chairman of the Undergraduate Program in Computer Science (2012-2017), and Rector's delegate for Education: postgraduates and Phd (2013-2016). He is currently the Scientific Coordinator of the PhD Program in Computer Science (XXXIII-XXXVI cycles) and Deputy Rector for Strategic Planning and Information Systems at the University of Catania. His research interests include Computer Vision, Imaging technology and Multimedia Forensics.

\end{biography}

\end{document}